# Improve Machine Learning carbon footprint using Nvidia GPU and Mixed Precision training for classification models

## Part I

**Andrew Antonopoulos**

andrew.antonopoulos@sony.com

## Abstract

This is the 1st part of the dissertation for my master's degree and compares the power consumption using the default floating point (32-bit) and Nvidia's mixed precision (16-bit and 32-bit) while training a classification ML model. A custom PC with specific hardware was built to perform the experiments, and different ML hyper-parameters, such as batch size, neurons, and epochs, were chosen to build Deep Neural Networks (DNN). Additionally, various software was used during the experiments to collect the power consumption data in Watts from the Graphics Processing Unit (GPU), Central Processing Unit (CPU), Random Access Memory (RAM) and manually from a wattmeter connected to the wall. A benchmarking test with default hyper-parameter values for the DNN was used as a reference, while the experiments used a combination of different settings. The results were recorded in Excel, and descriptive statistics were chosen to calculate the mean between the groups and compare them using graphs and tables. The outcome was positive when using mixed precision combined with specific hyper-parameters. Compared to the benchmarking, the optimisation for the classification reduced the power consumption between 7 and 10 Watts. Similarly, the carbon footprint is reduced because the calculation uses the same power consumption data. Still, a consideration is required when configuring hyper-parameters because it can negatively affect hardware performance. However, this research required inferential statistics, specifically ANOVA and T-test, to compare the relationship between the means. Furthermore, tests indicated no statistical significance of the relationship between the benchmarking and experiments. However, a more extensive implementation with a cluster of GPUs can increase the sample size significantly, as it is an essential factor and can change the outcome of the statistical analysis.

# 1 Introduction

The greenhouse effect is a natural phenomenon related to the sun's radiation, which travels towards the Earth [1]. The radiation reaches the earth and is absorbed by the land and ocean, and some are released toward space [1]. Most of it is captured and retained by greenhouse gases, a combination of chemical compounds that help keep Earth at a suitable temperature for all living beings [2]. Gases like carbon dioxide are produced naturally or by human activities, and by increasing it will also increase the Earth's temperature, affecting everyone's life [2]. The carbon footprint is the total amount of carbon dioxide emitted by human actions and is measured in grams of $CO_2$ (Carbon dioxide) equivalent per kilowatt hour (gCO2e/kWh) [3]. The higher the carbon footprint, the more impact it will have on the environment.

# 2 Background

Machine Learning (ML) has become very popular in many industries, and various services, such as cybersecurity, healthcare, and finance, have adopted it [4]. Millions of people use ML services hosted in the Cloud and specifically in big data centres [5]. This forces service providers to build big data centres to store the hardware and support growth. The data centres require cooling systems and power generators to maintain thousands of servers, consuming substantial power sources such as water and electricity [5]. Therefore, ML services are increasing and overloading many data centres worldwide, which can affect their sustainability, eventually increasing the carbon footprint and affecting the environment.

Data centres are using energy from non-fossil-fuelled technologies (solar, wind, hydro) instead of fossil-fuelled technologies (coal, oil, gas) [3]. However, there are no carbon-free forms of generating energy [3], and optimising ML services is a potential candidate to help reduce the carbon footprint.

# 3 Methodology

Figure 1 shows the steps followed to generate and collect data. Initially, the dataset, taken from Kaggle and included bird species in 84,635 training images, 2,625 test images, and 2,625 validation images, was used for classification. Various experiments were created by utilising different ML optimisation techniques and hyperparameters. The data were collected into an Excel file and used for analysis during the experiments. This procedure was repeated until it satisfied all the experiment use cases.

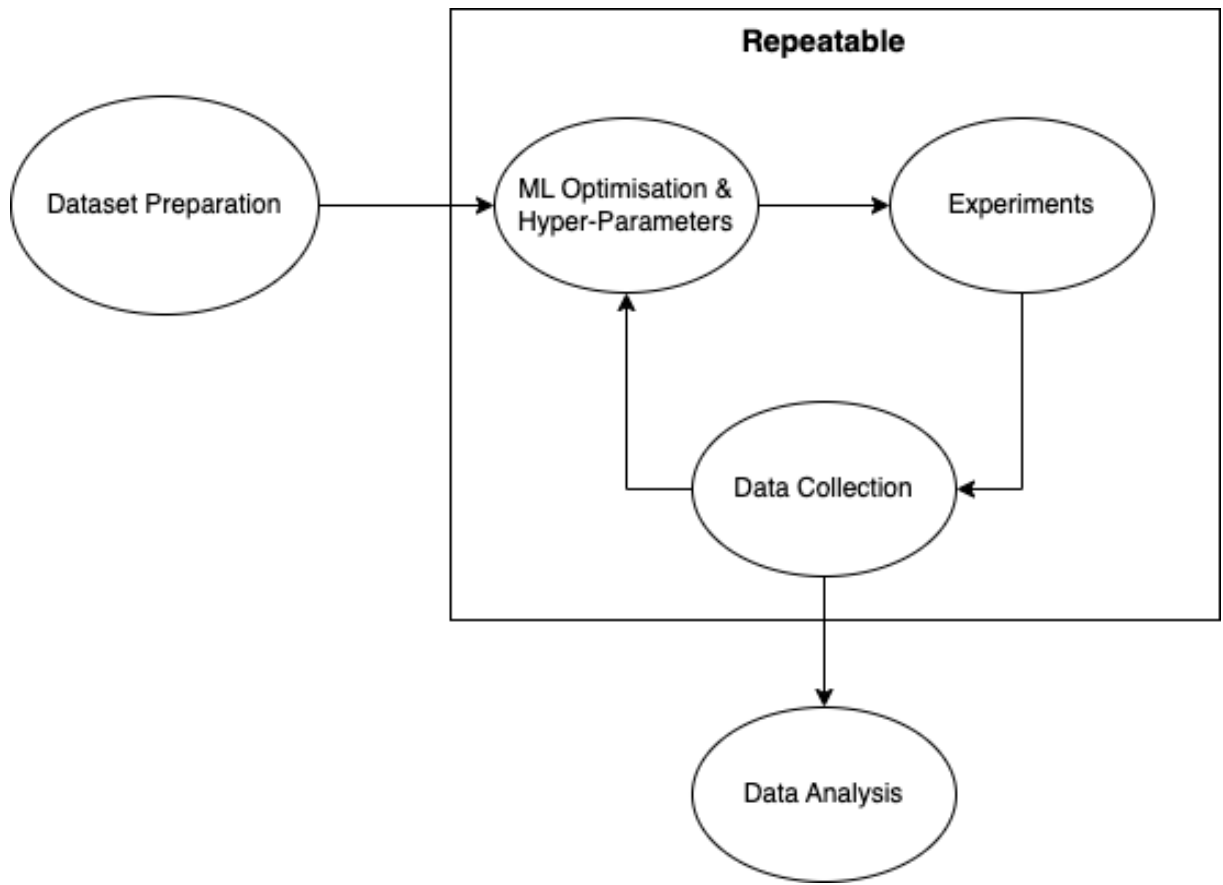

**Figure 1: Research steps during experiments and data collection**

A custom PC was built, which was used during the experiments to produce and collect the data. The hardware components were the following:

| Component (Hardware/Software) | Model |
|---|---|
| Motherboard | MSI Z690 DDR4 |
| CPU | Intel i5-12600 |
| Memory | Kingston Fury 32GB (4x8GB) |
| GPU | MSI NVIDIA RTX 4060 16GB GDDR6 (18Gbps/128bit) |
| SSD | Kingston A400 500GB |
| PSU | EVGA 1600w P2 |
| OS | Windows 10 Pro |

Choosing the appropriate ML framework is crucial, as it allows models to be developed without understanding the underlying algorithms. The most popular and common frameworks are TensorFlow and PyTorch [6]. TensorFlow performs better and is highly accurate on the coloured images dataset, while PyTorch is ideal for the black-and-white images dataset [7]. Furthermore, TensorFlow is more flexible and preferable for tasks that require high precision, with more control over the training flow [8].

PyTorch uses less GPU than it does with the CPU, unlike TensorFlow, which utilises the GPU more efficiently [9]. This is an essential advantage for TensorFlow because the author used GPU during the research's lifetime.

Besides GPU utilisation and accuracy, TensorFlow has better memory management than PyTorch, which is essential for large batch sizes and can improve power consumption [10]. Both frameworks allow users to adjust hyper-parameters to achieve better results, but TensorFlow can accomplish the same results with fewer lines of code and complexity [11].

TensorFlow requires experience; however, Keras, a high-level API that runs on top of TensorFlow, provides a quick implementation, has a simple architecture, and focuses on the user experience to accelerate the development of DNNs [12]. Therefore, TensorFlow and Keras were used to develop the ML models and perform the experiments.

## 3.1 Collecting computation power consumption data

Identifying the hardware and software to collect power consumption data is a crucial step. The GPU is responsible for around 70% of power consumption. In comparison, the CPU is responsible for 15%, RAM for 10%, and the remaining 5% from other PC components [13]. Therefore, the GPU, CPU and RAM are critical components because they directly impact the ML lifecycle. SSD or HDD are also crucial but are used by the operating system and other processes, so it is challenging to clarify the direct relationship to the ML process [14].

Furthermore, a comparison between the software that collects power consumption data identified that Code Carbon was more accurate [13]. However, it uses a fixed value for RAM and summarises the power consumption for all CPU cores [13]. Therefore, additional software was used to overcome Code Carbon limitations, which are listed below:

1. **Comet** automatically creates an Emissions Tracker object from the code carbon package to visualise the experiment's carbon footprint.

2. **Code Carbon v3.35.3** is lightweight software that seamlessly integrates into the Python codebase. It estimates the amount of carbon dioxide ($CO_2$) that the personal computing resources produce when executing the code.
3. **HWiNFO v7.66-5271**, focuses on hardware and categorises all the information it collects into sections. It can also collect power consumption for the CPU and GPU.
4. **Core Temp v1.18.1**, is a compact and powerful program for monitoring processor temperature and other vital information, such as power consumption.
5. **MSI Afterburner v4.6.5**, provides an on-screen display, hardware monitoring, custom fan profiles, and video capture. Additionally, it includes power consumption for the GPU and CPU.
6. **Corsair iCUE v5.9.105,** allows customisation of its various supported components and peripherals and provides information on how the GPU and CPU are used.
7. **Intel Power Gadget v3.6** is a software-based power estimation tool explicitly designed to monitor power consumption and utilisation for Intel Core processors.
8. **Wattmeter** was used to monitor the overall power consumption connected to the wall socket and the PC's power supply directly to the wattmeter.

## 3.2 ML optimisation techniques

Optimisation is crucial when creating a more efficient DNN because it has a certain level of complexity. Hyper-parameter optimisation techniques, such as the number of hidden layers, batch size, neurons, and epochs, cannot be modified individually and manually because they require a lot of time and experience [15]. If a non-optimal hyper-parameter is chosen for a particular reason, the DNN will consume more processing power [16]. The hyper-parameter will require fine-tuning to achieve the ideal results, but DNNs may fail to train or receive inefficient results because of the non-optimal values [17].

Researchers created a framework to adjust the hyperparameters dynamically to reduce time and computation resources during ML model training [18]. An alternative framework has also been developed to analyse GPU performance and find an optimal configuration to balance GPU power consumption and execution time [19]. Besides the benefit of automatically adjusting hyper-parameters and GPU configuration, we also need to consider the mixed precision technique, a new feature supported only by the latest GPUs. Mixed precision can use 32-bit and 16-bit floating points to represent the weights during the neural network training [20]. NVIDIA engineers claim that using mixed precision can help reduce the hardware's power consumption and keep the accuracy similar to a 32-bit floating-point method [20].

Therefore, this research utilised mixed precision and hyper-parameters to evaluate the benefit of creating an ML model from the power consumption perspective.
The chosen hyper-parameters are the following:

- **Neurons** determine the amount of information stored in the network, and more neurons allow us to learn more complex patterns. It can also increase the number of network connections, which requires more computational resources [21].
- **Batch size** is the number of training samples used to train a neural network. To fully take advantage of the GPU's processing, the batch size should be a power of 2 [22].
- **Epochs** are the number of complete passes of the training dataset through the algorithm's learning process, and the default values were identified during the pre-tests [15].

### 3.3 Power Consumption Data

Figure 2 shows the architecture and how data were collected. Multiple third-party software extracted the RAM, CPU, and GPU utilisation and power consumption data in Watts. The data were collected in an Excel file for comparison and generating the average value. The PSU was connected directly to the wattmeter, but because the software is unavailable, reading the values manually is required.

Code Carbon, a Python library, was integrated into the Python code, and data was seamlessly collected while the code was running. However, Code Carbon cannot store historical data, and Comet has been used to retrieve the average value over a period of time. Comet is a web service that pulls data from Code Carbon via an API to monitor GPU and CPU power consumption and utilisation. The collected data from all the software and the wattmeter was imported into Excel for further analysis.

Watts have been chosen because they measure the power consumed by a device. The higher the wattage, the more significant the amount of electrical power the PC uses over a period of time.

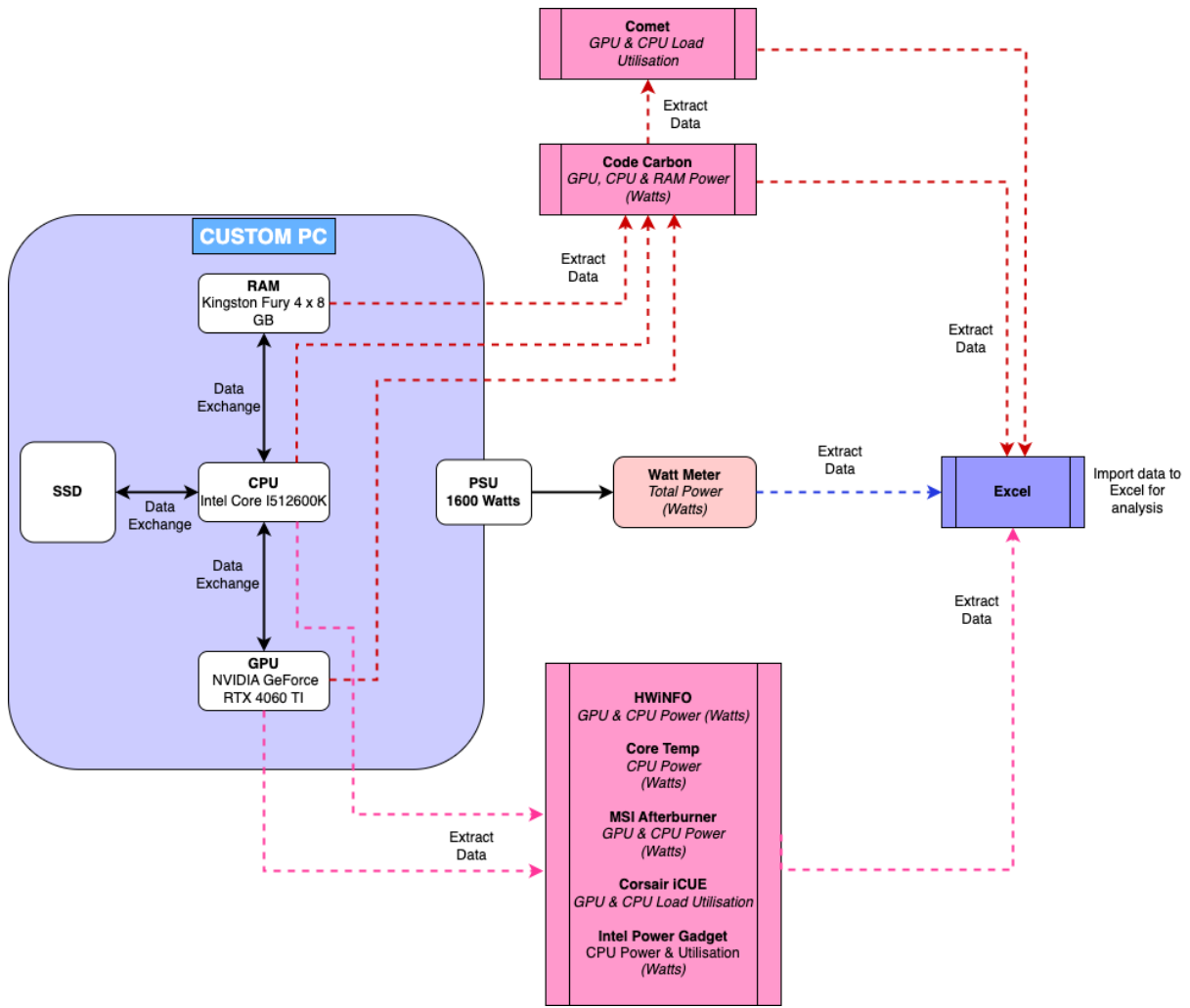

Figure 2: Overall architecture to collect power consumption data

### 3.4 Data Analysis Technique

After completing the experiments and the data extraction, descriptive statistics were adapted to assess the central tendency of the power consumption values. The author used a component bar chart to illustrate the comparison between the average of each piece of hardware [23]. However, further analysis of the findings using inferential statistics was required because the differences between the average values were too close. To achieve this task, ANOVA was used to evaluate the relationship between the tests and multiple T-tests were used to check whether the difference between experiments was statistically significant [24]. Figure 3 summarises the steps that followed during the analysis.

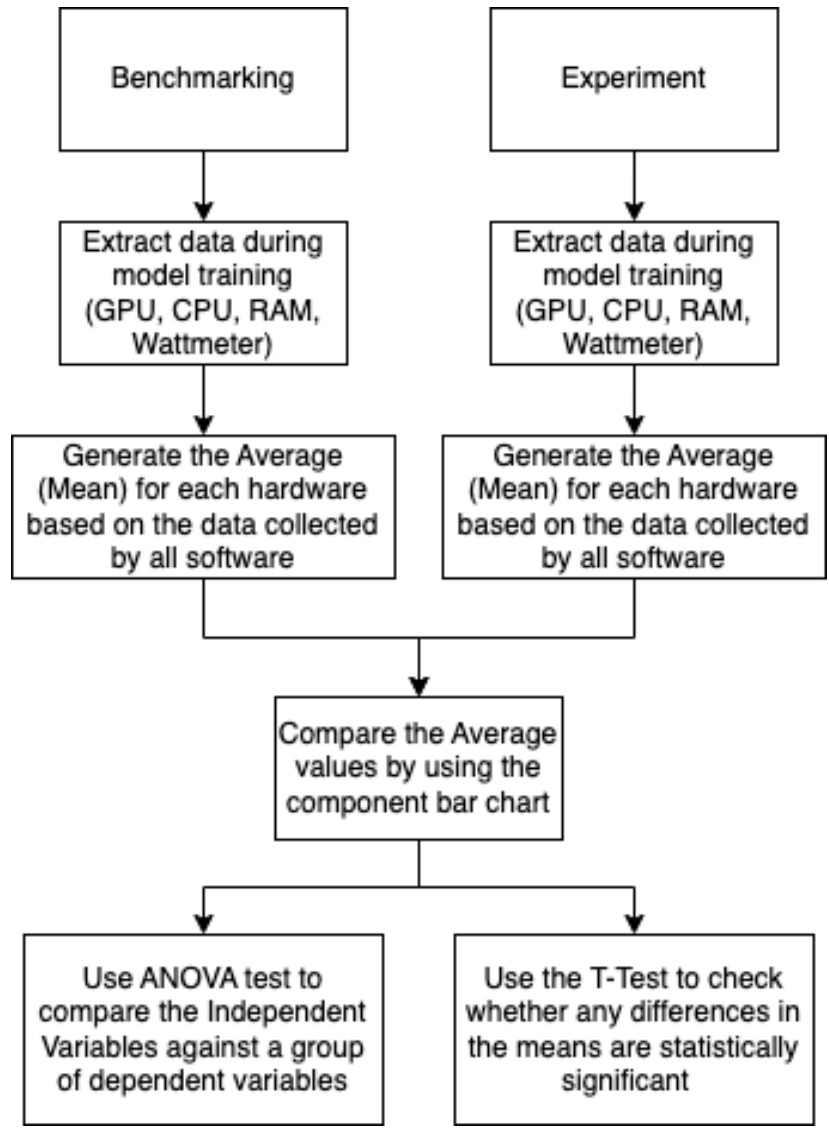

Figure 3: Steps that followed during the data analysis

## 4 Testing and Results

### 4.1 Introduction

The GPU has played a vital role in ML and model training because it is powered by Tensor Cores, which are specialised cores that enable mixed precision and can accelerate training and learning performance [25]. Using a GPU that supports Tensor Cores, we can utilise the mixed precision functionality, accelerating the throughput and reducing AI training times [25].
This research used the NVIDIA RTX 4060 Ti Ventus, which supports overclocking and operates at 2595 MHz instead of 2565 MHz in standard mode. This frequency indicates how much data it can process per clock cycle. Additionally, it supports the 4[th] generation of NVIDIA's Tensor Cores and the latest technology in high-performance memory GDDR6 with a capacity of 16 GB. However, the most important is the high memory bandwidth of 18 Gbps, which allows fast data transfer between the GPU memory and the computation cores.

To use the mixed precision, the libraries have been imported into the Python code and configured to be used with the public policy. After the implementation and execution of the code, the mixed precision library checked the GPU and reported the version of the computation capability. The computation capability identifies the features supported by the GPU hardware and is used by applications at runtime to determine which hardware instructions are available [20]. According to the mixed precision Python library, the compute capability version must be more than 7.0. The GPU that has been used for this research have a compute capability version of 8.9, as Figure 4 shows:

```
INFO:tensorflow:Mixed precision compatibility check (mixed_float16): OK
Your GPU will likely run quickly with dtype policy mixed_float16 as it has comp
ute capability of at least 7.0. Your GPU: NVIDIA GeForce RTX 4060 Ti, compute c
apability 8.9
Compute dtype: float16
Variable dtype: float32
```

**Figure 4: Mixed precision and compute capability reported by the Python library**

The above output indicates that the current GPU will use a floating point of 16-bit for computations to improve performance and 32-bit for the variables, mainly for numerical stability, so the model trains with the same quality.

During the benchmarking for the classification model training, the default floating point of 32-bit was used, while all the experiments used only mixed precision.

## 4.2 Classification

As shown in Figure 5, a series of steps were followed to collect the data during the benchmarking and experiments.

The dataset includes 89,885 images of 525 species. Each image has a width and height of 224 X 224 pixels, respectively, and the depth is 3 because it is an RGB image in JPG format. However, the photos are of high quality and didn't require any modification.

After loading the dataset, the floating point was configured with the associated hyper-parameters such as batch size, number of neurons, and epochs. Measurements were taken before and during the model training and stored separately to be used for the comparison. In addition to the software-based data collection, the wattmeter data were taken visually by checking the equipment every fifteen minutes and manually calculating an average.

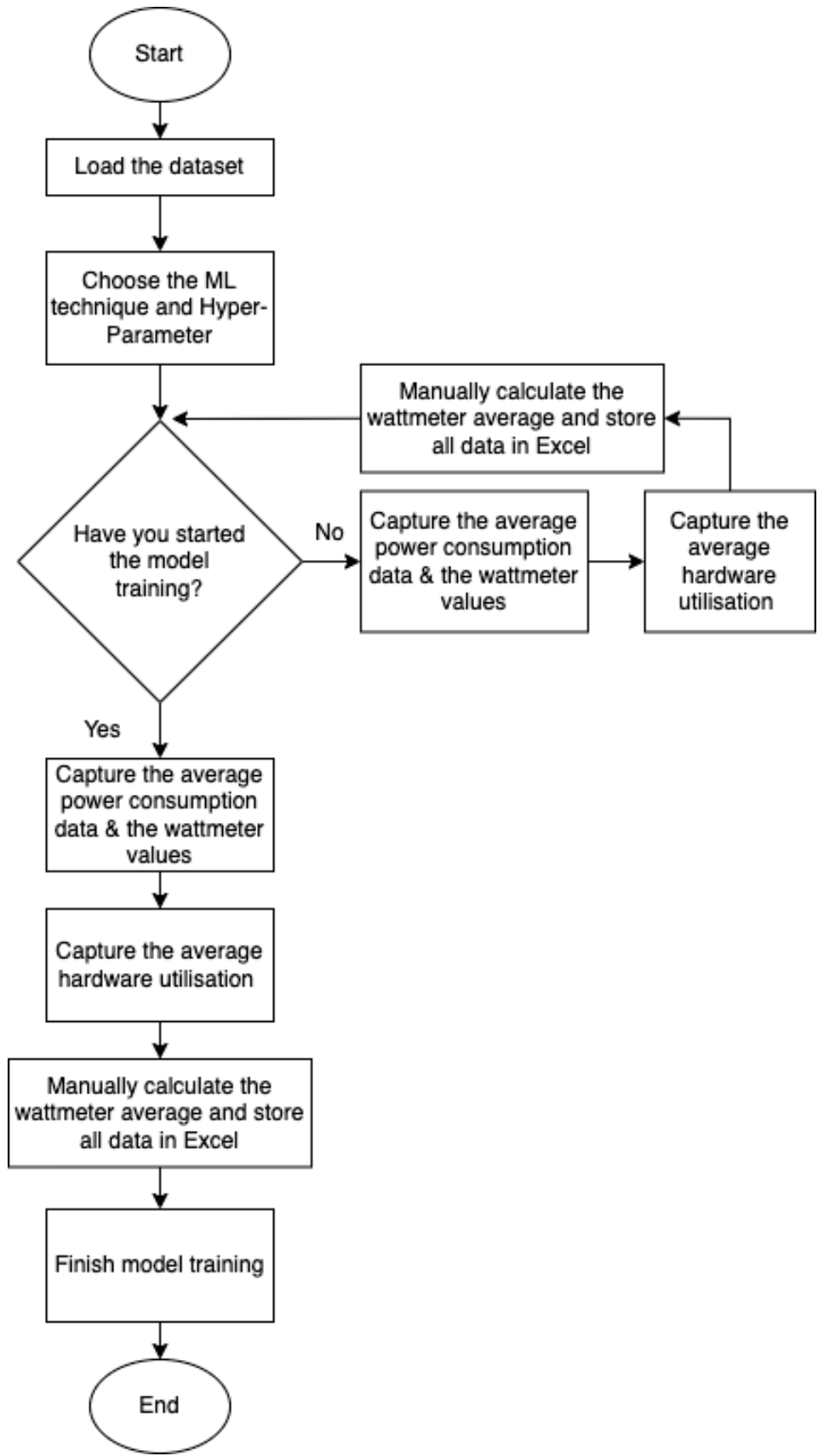

**Figure 5: Flow of the classification testing and data collection**

## 4.3 Benchmarking

A DNN has been created for the benchmarking test with the hyper-parameters shown in Table 1.

|  | Benchmarking |
|---|---|
| Floating Point | 32 |
| Batch Size | 32 |
| Neurons | 1024 |
| Epochs | 25 |

**Table 1: Hyper-parameters for the classification benchmarking**

Figure 6 shows the power consumption before and during the model training. Initially, the GPU's power consumption was 8 Watts, the CPU 16 Watts, and the RAM 12 Watts. However, the overall power consumption was 70 Watts because of the power usage from other hardware components, such as fans, SSD, and motherboard. To keep the PC cool, the author used 9 fans, two dedicated to the GPU, two to the CPU, and the rest for internal airflow.

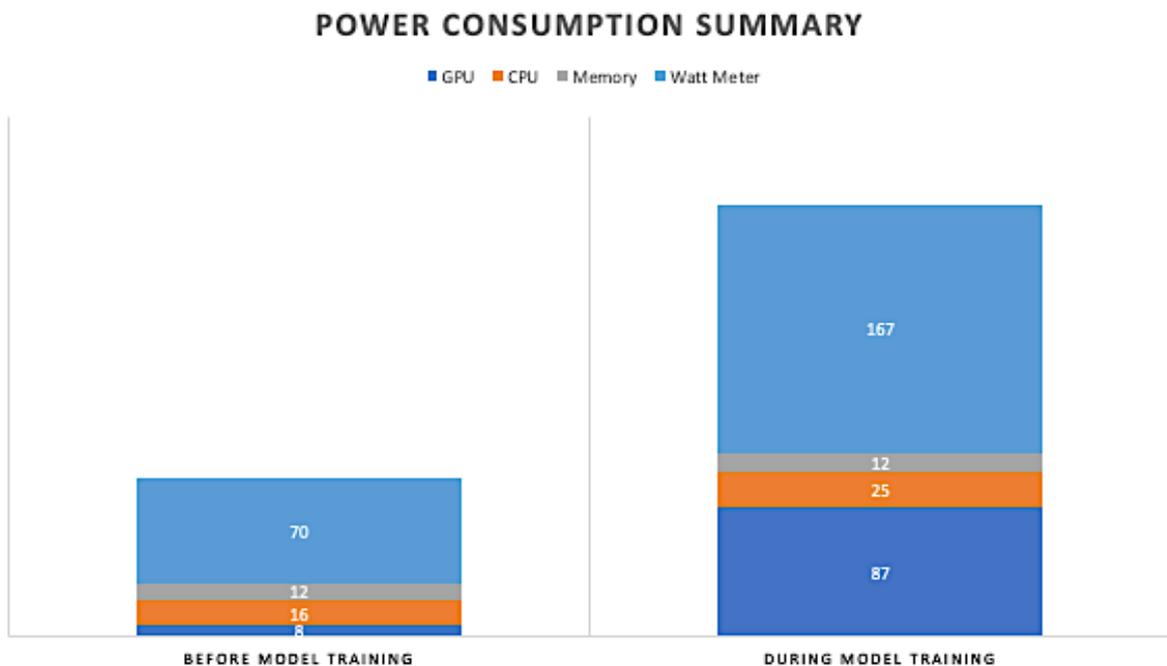

**Figure 6: Power consumption data for classification benchmarking**

During the model training, GPU power consumption increased to 87 Watts, CPU to 25 Watts, and RAM to 12 Watts because of the fixed value provided by Code Carbon. However, the wattmeter reported that the total power consumption increased to 167 Watts.

Similarly, the GPU and CPU utilisation was 2% before the model training and increased to 93% for the GPU and 7% for the CPU, indicating that the GPU was utilised during the model training. Unfortunately, none of the software reported RAM utilisation, as shown in Figure 7.

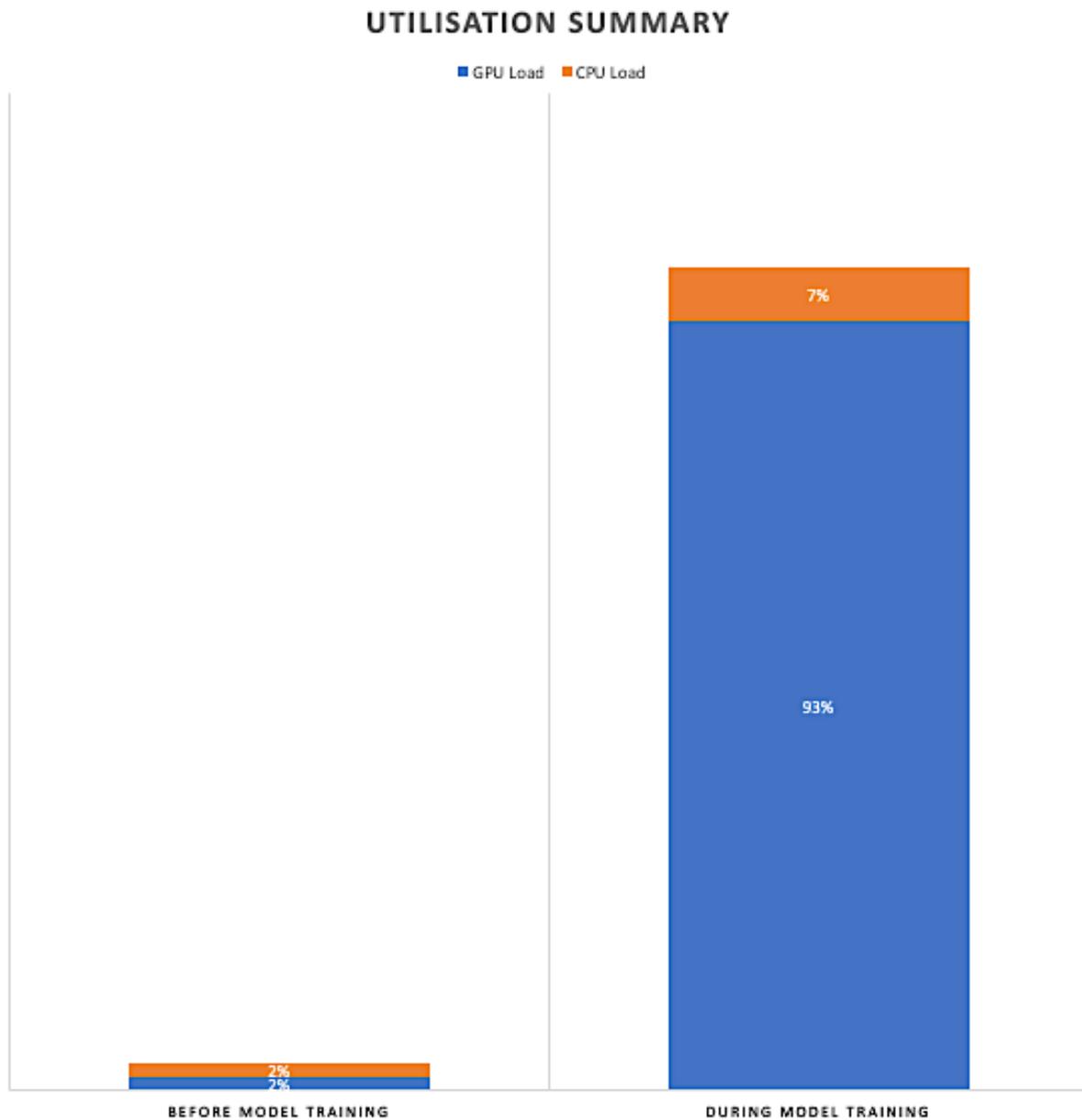

Figure 7: Utilisation data for classification benchmarking

## 4.4 Experiments

For each experiment, batches and neuron values differed from the benchmarking results. Table 2 shows the configurations for the DNN networks. The common factor is mixed precision and epoch values because they can affect the duration of the model. Therefore, the author kept the same value to avoid affecting the model training duration between the benchmarking and the experiments.

|  | 1st Experiment | 2nd Experiment | 3rd Experiment |
|---|---|---|---|
| Floating Point | Mixed Precision | Mixed Precision | Mixed Precision |
| Batch Size | 32 | 256 | 256 |
| Neurons | 1024 | 1024 | 2048 |
| Epochs | 25 | 25 | 25 |

Table 2: Hyper-parameters for the classification experiments

The same steps as the benchmarking testing were followed and power consumption with the utilisation data were collected before and during the model training. Figure 8 compares the wattage power consumption before the model training, and Figure 17 shows the hardware utilisation during the same period. These results indicate that before the model training, the PC was idle, and no unnecessary processes were running in the background, which could affect the power consumption and hardware utilisation during the tests.

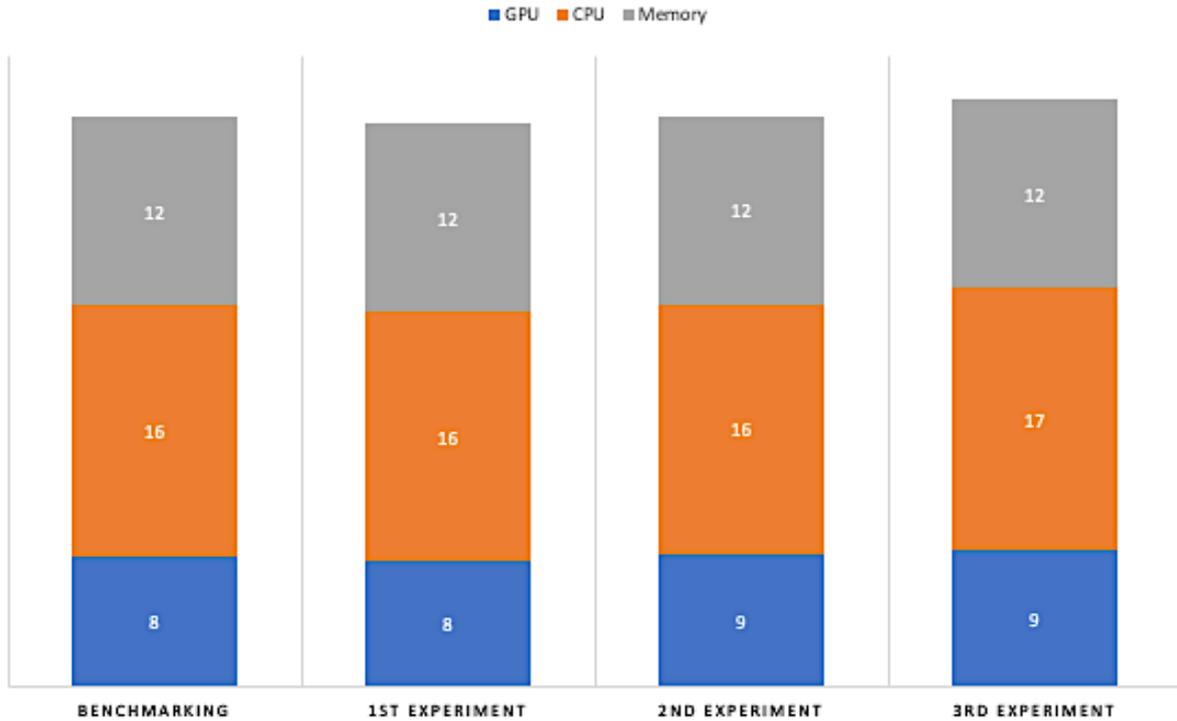

Figure 8: Power consumption data before the model training

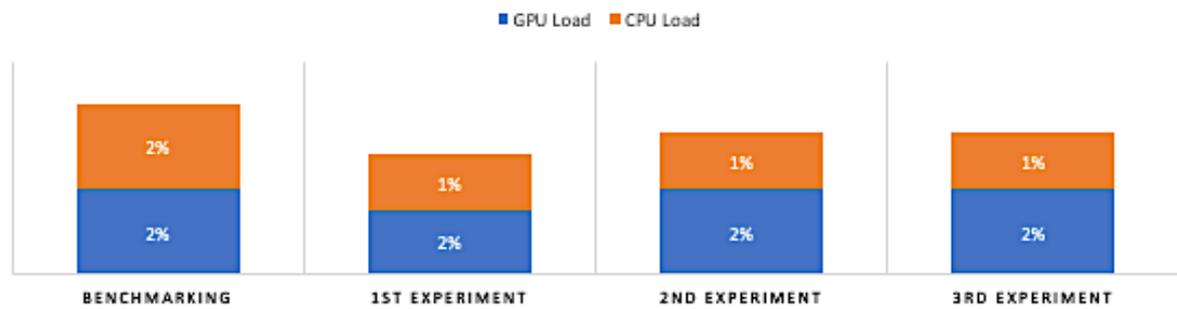

Figure 9: Hardware utilisation data before the model training

The total power consumption, taken from the wattmeter, reported similar values across the tests, as shown in Figure 10. This data confirms that the PC's operation before the model was normal, and similar workloads were used.

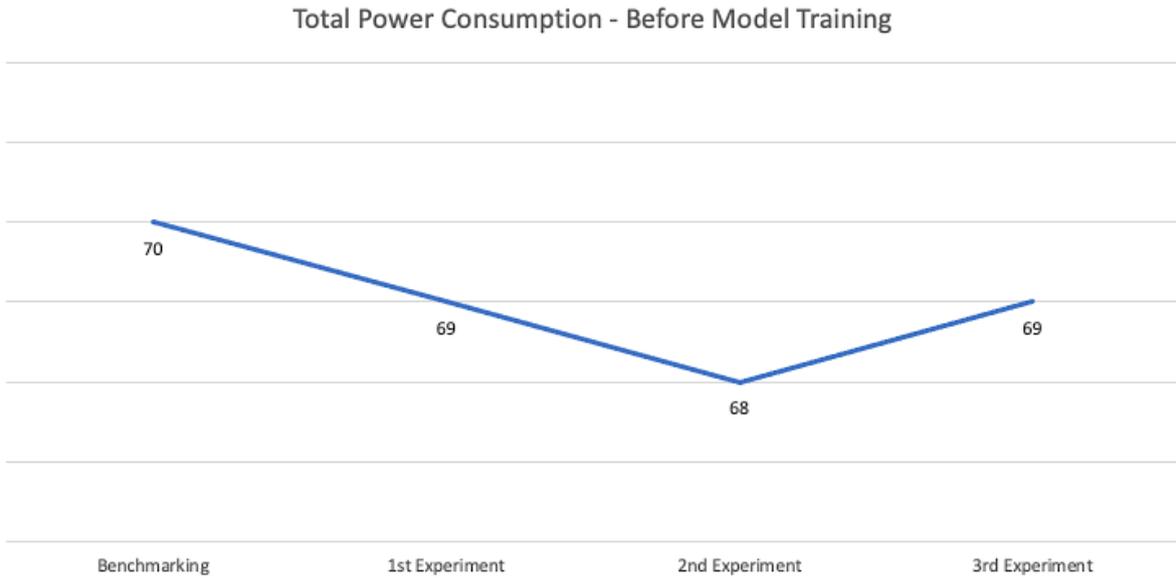

**Figure 10: Total power consumption before model training between tests**

During the model training, the GPU power consumption increased but, as shown in Figure 11, dropped by 3 Watts in the 1st experiment, 7 Watts in the 2nd experiment and 6 Watts in the 3rd experiment compared to the benchmarking data.

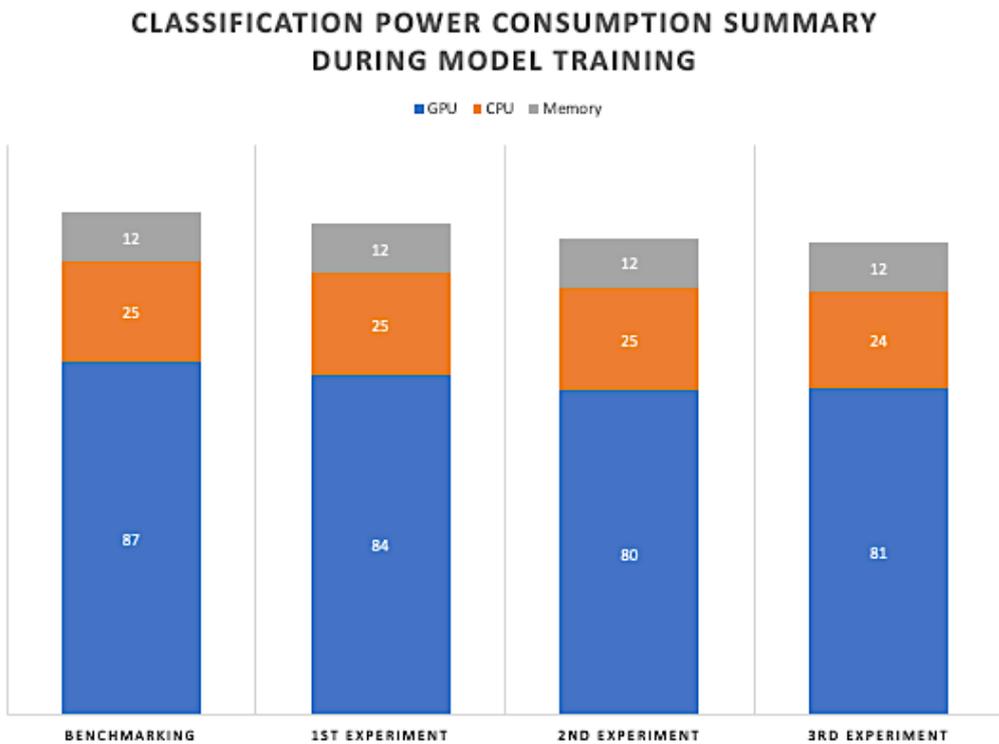

**Figure 11: Power consumption data during the model training**

The GPU utilisation decreased by 3% in the 1st experiment and 2% in the 2nd experiment but increased by 2% in the 3rd experiment compared to the benchmarking data, as shown in Figure 12.

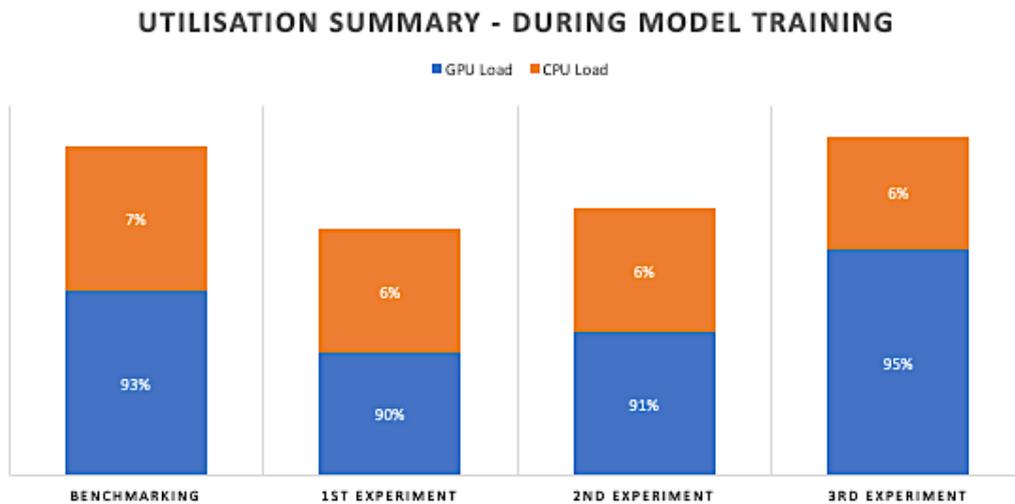

**Figure 12: Hardware utilisation data during the model training**

Overall, Figure 13 shows that the power consumption during the model training decreased by 7 Watts for the 1st and 2nd experiments and 10 Watts for the 3rd experiment compared to the benchmarking data.

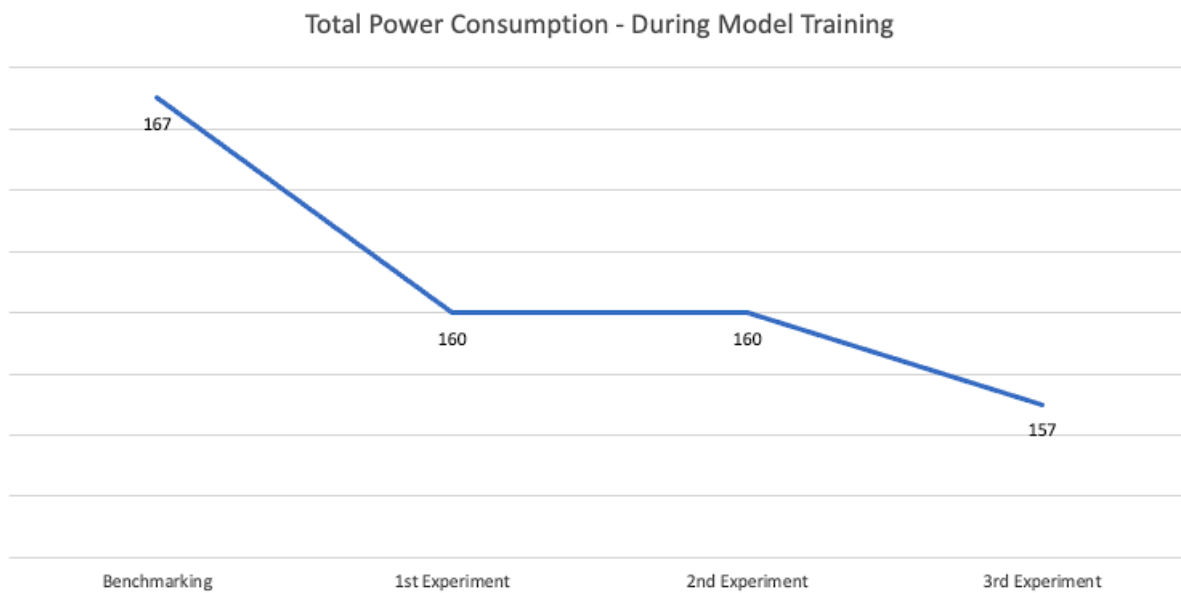

**Figure 13: Total power consumption during the model training between tests**

Figure 13 indicates that the mixed precision (MP) helped reduce the GPU and overall power consumption. The batch size, which was increased in the 2nd and 3rd experiments, helped slightly improve the power consumption compared to the 1st experiment. However, the 3rd experiment, which had the double number of neurons and forced the GPU to work harder (an extra 2%), dropped the overall power consumption by 10 Watts compared to the benchmarking and 3 Watts compared to the 1st and 2nd experiments.

Calculating the carbon footprint for the classification tests will require four steps, as shown in Table 3 [26]. The 1st step is to use the overall power consumption from Figure 13 and convert it to kilowatts (kW) using equation (1).

$$kW = Watts\ /\ 1000 \tag{1}$$

The 2nd step is to convert the kW to Kilowatt per hour (kWh) by using equation (2) and the duration of the model training, which was 2 hours for each classification test.

$$kWh = kW * Number\ of\ hours\ for\ model\ training \tag{2}$$

The 3rd step is to collect the carbon intensity from a public website [27] for the specific day the tests took place. The last step is calculating the carbon footprint using equation (3).

$$Carbon\ Footprint\ (gCO2e/kWh)$$
$$= Power\ Consumption\ (kWh) * Carbon\ Intensity\ (gCO2e) \tag{3}$$

The outcome of the carbon footprint results is similar to power consumption and the following table showing that the 3rd experiment produced less emissions among all.

| Steps | Benchmarking | 1st Experiment | 2nd Experiment | 3rd Experiment |
|---|---|---|---|---|
| Power Consumption | 167 Watts | 160 Watts | 160 Watts | 157 Watts |
| Step 1 : kW | 0.167 kW | 0.16 kW | 0.16 kW | 0.157 kW |
| Step 2 : kWh | 0.334 kWh | 0.32 kWh | 0.32 kWh | 0.314 kWh |
| Step 3 : Carbon Intensity | 268 gCO2e | 268 gCO2e | 268 gCO2e | 268 gCO2e |
| Step 4 : Carbon Footprint | 89.512 gCO2e/kWh | 85.76 gCO2e/kWh | 85.76 gCO2e/kWh | 84.152 gCO2e/kWh |

**Table 3: Classification of carbon footprint calculation**

Overall, mixed precision is essential in reducing power consumption. The results in Table 3 confirm that hyper-parameters require time and experience to find the ideal hyper-parameters because they can affect the hardware performance positively or negatively. In this case, adjusting the batch size and neurons further improved the carbon footprint as per the 3$^{rd}$ experiment.

# 5   Analysis and Evaluation

## 5.1   Introduction

During the analysis, four groups were used to identify a potential statistical significance based on their means using the ANOVA test, as shown in Figure 14. Each group has four values: GPU, CPU, RAM and total power consumption, which were taken from the Wattmeter.
ANOVA can be used when we have more than two groups, but if there is a significant difference, it does not illustrate where the significance lies [28]. Therefore, multiple T-tests have been used to compare the means between a combination of two groups [28].

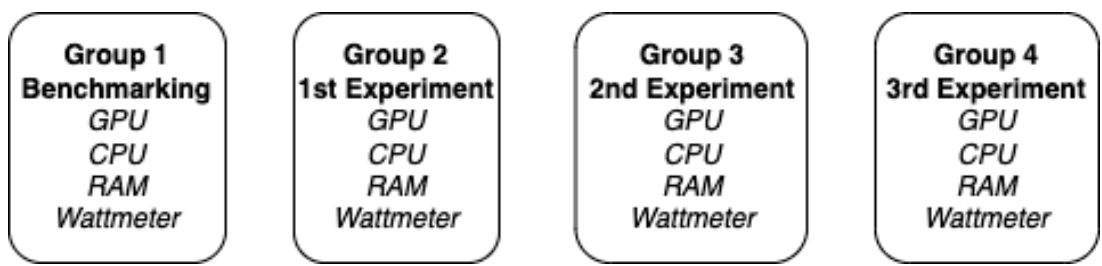

**Figure 14: Groups used for the inferential analysis**

## 5.2   Classification Analysis

To be able to use ANOVA for the classification data will require to fulfil the following assumptions [29]:

1. The data in each group are normally distributed
2. The data in each group have the same variance
3. The data are independent

The data collected during each classification test is summarised in Table 4.

|  | Benchmarking<br>Floating Point: 32<br>Batch Size: 32<br>Neurons: 1024<br>Epochs: 25 | 1st Experiment<br>Floating Point: MP<br>Batch Size: 32<br>Neurons: 1024<br>Epochs: 25 | 2nd Experiment<br>Floating Point: MP<br>Batch Size: 256<br>Neurons: 1024<br>Epochs: 25 | 3rd Experiment<br>Floating Point: MP<br>Batch Size: 256<br>Neurons: 2048<br>Epochs: 25 |
|---|---|---|---|---|
| GPU | 87 Watts | 84 Watts | 80 Watts | 81 Watts |
| CPU | 25 Watts | 25 Watts | 25 Watts | 24 Watts |
| RAM | 12 Watts | 12 Watts | 12 Watts | 12 Watts |
| Wattmeter | 167 Watts | 160 Watts | 160 Watts | 157 Watts |

**Table 4: Power consumption data during model training**

Skewness and Kurtosis were calculated in Excel using descriptive statistics to validate the distribution normality. The accepted values for Skewness are between -2 and +2, while the kurtosis value is between -7 and +7 [30]. Table 5 shows the values for all the tests are within the suggested range, indicating that the distribution is normal.

|  | Benchmarking | 1st Experiment | 2nd Experiment | 3rd Experiments |
|---|---|---|---|---|
| Kurtosis | -0.57792078 | -0.515980918 | -0.060450943 | -0.382528412 |
| Skewness | 0.950385095 | 0.953821581 | 1.043778768 | 0.990587468 |

**Table 5: Kurtosis and Skewness values between tests**

The F-test was used in Excel to calculate the F-value between the two groups and compare the variance. Six tests have been completed, as shown in Table 6.

|  | Benchmarking | 1st Experiment | 2nd Experiment |
|---|---|---|---|
| 1st Experiment | 1.105609376 |  |  |
| 2nd Experiment | 1.114340483 | 1.007897099 |  |
| 3rd Experiment | 1.144927028 | 1.035561974 | 1.027448114 |

**Table 6: F-Test variance comparison between tests**

If the value is closer to 1.5 or less, the sample variance is equal, and we can confidently perform the ANOVA test [31]. Therefore, Table 6 indicates that variance is equal and that ANOVA can be used to validate the equality between the means [31]. Also, the data are independent because different use cases are being tested between the groups, which validates the third assumption.

The significant level (denoted as Alpha) that has been chosen is 0.05 (or 5%), which means that there is a 95% probability that the results found in the study are the results of a true relationship between the compared groups [32]. The values needed for the analysis are the P-value and the comparison between the F-Test and F-critical values as shown in the Table 7. If the P-value exceeds 0.05, we accept the null hypothesis ($H_0$) and reject the alternative hypothesis ($H_1$). If the F-value exceeds the F-critical for the selected Alpha level (0.05), we can reject the $H_0$ and accept the $H_1$ [33].

| Source of Variation | SS | Df | MS | F-value | P-value | F-critical |
|---|---|---|---|---|---|---|
| Between Groups | 41.99546875 | 3 | 13.99848958 | 0.00302786 | 0.999756527 | 3.490294819 |
| Within Groups | 55478.73688 | 12 | 4623.228073 |  |  |  |
| Total | 55520.73234 | 15 |  |  |  |  |

**Table 7: ANOVA Single Factor P-value calculation**

The average, shown in Table 8, indicates that the 3rd experiment has the preferable ML hyper-parameters to reduce the power consumption; however, the P-value is higher than 0.05, which suggests accepting $H_0$, and the F-value is smaller than the F-critical, which also suggests accepting $H_0$.

$H_1$ predicts a difference between the groups, meaning that changes applied during the experiments can improve power consumption. In contrast, $H_0$ always contains the condition of equality or no relationship between the groups [32]. In other words, changes to ML optimisation techniques will not improve the power consumption.

| Groups | Count | Sum | Average (Mean) | Variance |
|---|---|---|---|---|
| Benchmarking | 4 | 290.6 | 72.65 | 5031.69 |
| 1st Experiment | 4 | 280.7 | 70.175 | 4551.055833 |
| 2nd Experiment | 4 | 277.15 | 69.2875 | 4515.397292 |
| 3rd Experiment | 4 | 273.1 | 68.275 | 4394.769167 |

Table 8: ANOVA Single Factor summary for each group

On the other hand, two types of errors need to be considered: Type I and Type II. Type I errors occur when the researcher incorrectly rejects a true $H_0$, while Type II errors occur when the researcher accepts the $H_0$ when it should be rejected [32].

To reduce the chances of making Type I and Type II errors requires adjusting the P-value. A small P-value can increase Type II errors, while a high P-value (0.05 or above) can increase Type I errors, which are more critical for this research; therefore, a smaller P-value is preferable [33]. The author adopted the Bonferroni correction by dividing the P-value (0.05) by the statistical tests performed [34].

Six paired T-tests were chosen in the Excel file to compare the means by using the Bonferroni correction:

$$New\ P-value = \frac{Old\ P-value}{Number\ of\ tests} = \frac{0.05}{6} = \mathbf{0.00833} \quad (4)$$

Paired T-tests can be used when the objects between the samples are related, for example, by comparing before and after results [35]. This is similar to the comparison between the classification tests because using the same equipment in different configurations. Additionally, H$_1$ is specific and unidirectional, as shown by the comparison of benchmarking with the experiment results. Therefore, one-sided P-values have been used to test H$_0$ [36], as shown in Table 9.

|  | 1st Experiment | 2nd Experiment | 3rd Experiment |
|---|---|---|---|
| Benchmarking | 0.127005262 | 0.110240813 | 0.080243545 |
| 1st Experiment |  | 0.210707629 | 0.038152 |
| 2nd Experiment |  |  | 0.1599891 |

Table 9: One-side paired T-Test P-values comparison

Comparing the T-Test P-values with the chosen P-value of 0.00833, as explained in equation (4), we can conclude that there is no significant difference between the means.
Both ANOVA and T-test reported higher calculated P-values than the predefined. Therefore, the conclusion is to accept the H$_0$ and reject H$_1$.

## 5.3 Limitations of the Analysis

The propositions might not be valid because the analysis has limitations. To begin with, the RAM power consumption is based on a fixed value provided by the software. The CPU and GPU measurements varied between the software and required using the average value. However, by monitoring the utilisation, the author could verify the GPU usage across all the tests. Additionally, manually calculating the average for the Wattmeter values was required.
A significant limitation is the sample size, which included a single GPU, CPU, RAM and Wattmeter. Therefore, if there is a slight difference in the relationship between variables or groups, as in this research, a large sample size will help obtain a more accurate statistic test [32].

# 6  Conclusion and Future Work

In this research, the author discussed the potential improvement of the ML carbon footprint by investigating different ML optimisation techniques. Current literature suggests that using mixed precision during the ML model training can improve the GPU's computation and performance [20]. Additionally, it requires using hyper-parameters, which are essential for creating DNN networks [15], but it is important to configure the hyper-parameters appropriately to avoid negatively affecting the GPU performance [17].

As the literature suggested [14], various software must be used to monitor the hardware utilisation and computation power consumption because of existing limitations.

The author used the benchmarking test as a reference point using default DNN parameters and then completed a series of experiments using the literature suggestions. Initially, the results from each classification experiment were compared with the associated benchmarking results using descriptive analysis, specifically the mean. The classification results show that using mixed precision with more neurons can improve power consumption.

After summarising the test results, the author analysed the data using inferential statistics, specifically ANOVA and T-test. The commonality between the two tests is that both compare the means between groups. Still, ANOVA has a different way of determining the statistical significance and can be used when groups are more than three. Therefore, the author used ANOVA to compare the benchmarking with the three experiments. After the ANOVA comparison, the author used a T-test to compare multiple pairs of groups to cross-validate the ANOVA results and reduce Type I and Type II errors. The results reported no statistical significance between the means in classification and accepted $H_0$.

However, some limitations can affect the generalisation of this research, which future studies can probably overcome. First, future research could use a single software that supports a wide range of hardware to collect power consumption data more accurately and frequently. Second, future research may use a larger implementation with a cluster of GPUs, which will help to increase the sample size significantly because it is an essential factor for statistical analysis and can affect the outcome [32].